\documentclass[letterpaper]{article}
\usepackage[margin=1in,dvips]{geometry}
\usepackage{graphicx,psfrag,amsmath,amssymb}
\usepackage[numbers,sort&compress]{natbib}
\usepackage{url,color,booktabs}
\usepackage{array}
\usepackage{multirow}
\usepackage{enumitem}
\usepackage{listings}
\usepackage{hyperref}

\definecolor{DarkRed}{rgb}{0.545,0,0}
\hypersetup{
  breaklinks=true, colorlinks=true, linkcolor=black,
  citecolor=black, urlcolor=DarkRed,
  pdftitle={A flexible, extensible software framework for model compression based on the LC algorithm},
  pdfauthor={ Yerlan Idelbayev and Miguel {\'A}.\ Carreira-Perpi{\~n}{\'a}n},
}

\definecolor{mygreen}{rgb}{0,0.6,0}
\definecolor{mygray}{rgb}{0.5,0.5,0.5}
\definecolor{mymauve}{rgb}{0.58,0,0.82}

\def \ifempty#1{\def\temp{#1} \ifx\temp\empty }

\newcommand{\U}{\ensuremath{\mathbf{U}}}
\newcommand{\V}{\ensuremath{\mathbf{V}}}
\newcommand{\W}{\ensuremath{\mathbf{W}}}

\newcommand{\w}{\ensuremath{\mathbf{w}}}

\newcommand{\z}{\ensuremath{\mathbf{z}}}
\newcommand{\0}{\ensuremath{\mathbf{0}}}

\newcommand{\blambda}{\ensuremath{\boldsymbol{\lambda}}}

\newcommand{\btheta}{\ensuremath{\boldsymbol{\theta}}}

\newcommand{\bDelta}{\ensuremath{\boldsymbol{\Delta}}}

\newcommand{\bPi}{\ensuremath{\boldsymbol{\Pi}}}

\newcommand{\bTheta}{\ensuremath{\boldsymbol{\Theta}}}

\newcommand{\bbR}{\ensuremath{\mathbb{R}}}

\newcommand{\calC}{\ensuremath{\mathcal{C}}}

\newcommand{\calO}{\ensuremath{\mathcal{O}}}

\newcommand{\norm}[2][]{%
  \ifempty{#1} {\left\lVert#2\right\rVert} \else {#1\lVert#2#1\rVert} \fi}

\newcommand{\caja}[4][1]{{%
    \renewcommand{\arraystretch}{#1}%
    \begin{tabular}[#2]{@{}#3@{}}%
      #4%
    \end{tabular}%
    }}

{%
\begin{list}{#1}{
\vspace{-\topsep}
\vspace{-\partopsep}
\setlength{\itemindent}{0cm}
\setlength{\rightmargin}{0cm}
\setlength{\listparindent}{0cm}
\settowidth{\labelwidth}{#1}
\setlength{\leftmargin}{\labelwidth}
\addtolength{\leftmargin}{\labelsep}
\setlength{\itemsep}{0cm}
}%
}%
{%
\end{list}
\vspace{-\topsep}
\vspace{-\partopsep}
}

{\begin{enumerate}%
}%
{\end{enumerate}}

\DeclareMathOperator*{\argmin}{arg\,min}

\newcommand{\rankop}{\operatorname{rank}}
\newcommand{\rank}[1]{\ensuremath{\rankop\left(#1\right)}}

\newcommand{\colemph}[1]{\mbox{}{\color{black} #1}}
\newcommand{\colbf}[1]{\mbox{}{\color{magenta} #1}}
\newcommand{\colit}[1]{\mbox{}{\color{blue} #1}}

\graphicspath{{grf/},{grf-miguel/}}

\AtBeginDvi{}

\title{A flexible, extensible software framework \\ for model compression based on the LC algorithm}
\author{
  Yerlan Idelbayev and Miguel {\'A}.\ Carreira-Perpi{\~n}{\'a}n \\
  Department of CSE, University of California, Merced \\
  {\url{http://eecs.ucmerced.edu}} \\
  {\url{https://github.com/UCMerced-ML/LC-model-compression}}
}

\date{May 15, 2020}

\begin{document}

\maketitle

\begin{abstract}

  We propose a software framework based on the ideas of the \emph{Learning-Compression (LC) algorithm} \cite{Carreir17a,CarreirIdelbay17a,CarreirIdelbay18a, IdelbayCarreir20a,IdelbayCarreir20c}, that allows a user to compress a neural network or other machine learning model using different compression schemes with minimal effort. Currently, the supported compressions  include pruning, quantization, low-rank methods (including automatically learning the layer ranks), and combinations of those, and the user can choose different compression types for different parts of a neural network.
  
  The LC algorithm alternates two types of steps until convergence: a \emph{learning (L) step}, which trains a model on a dataset (using an algorithm such as SGD); and a \emph{compression (C) step}, which compresses the model parameters (using a compression scheme such as low-rank or quantization). This decoupling of the ``machine learning'' aspect from the ``signal compression'' aspect means that changing the model or the compression type amounts to calling the corresponding subroutine in the L or C step, respectively. The library fully supports this by design, which makes it flexible and extensible. This does not come at the expense of performance: the runtime needed to compress a model is comparable to that of training the model in the first place; and the compressed model is competitive in terms of prediction accuracy and compression ratio with other algorithms (which are often specialized for specific models or compression schemes). The library is written in Python and PyTorch and available in Github.

\end{abstract}

\section{Introduction}

With the success of neural networks in solving practical problems in various fields, there has been an emergence of research in neural network compression techniques that allows compressing these large models in terms of memory, computation, and power requirements. At present, many ad-hoc solutions have been proposed that typically solve \emph{only one specific type of compression}: quantization \cite{HwangSung14a, Courbar_15a, Rasteg_16a, Zhou_16b, Zhu_17a, Gong_15a, CarreirIdelbay17a}, pruning \cite{Lecun_90a, HassibStork93a, Han_15a, Liu_15a, Wen_16a}, low-rank decomposition \cite{Sainat_13a,Xue_13a,Denil_13a,Denton_14a,Jaderb_14a,Zhang_16c,Tai_16a,Wen_17a,LiShi18a,Xu_18a} or tensor factorizations \cite{Denton_14a,Lebedev_15a,Novikov_15a,Garipov_16a}, and others.

Among the various research strands in neural net compression, in our view, the fundamental problem is that in practice, one does not know what type of compression (or combination of compression types) may be the best for a given network. In principle, it may be possible to try different existing algorithms, assuming one can find an implementation for them, but practically it is often impossible. We seek a solution that directly addresses this problem and allows non-expert end-users to compress models easily and efficiently. Our approach is based on a recently proposed compression framework, the LC algorithm \cite{Carreir17a,CarreirIdelbay17a,CarreirIdelbay18a, IdelbayCarreir20a, IdelbayCarreir20c}, that by design separates the ``learning'' part of the problem, which involves the dataset, neural net model, and loss function from the ``compression'' part, which defines how the network parameters will be compressed. This separation has the advantage of \emph{modularity}: we can change the compression type by simply calling a different compression routine (e.g., $k$-means instead of the SVD), with no other changes to the algorithm. 

In this paper, we further develop the ideas of modular compression presented by LC algorithm and describe our ongoing efforts in building a software library with the philosophy of \emph{single algorithm --- multiple compressions}. At present, this handles 1) various forms of quantization, pruning, low-rank methods, and their combinations, 2) different types of deep net models, and 3) allows flexible configuration of compressed schemes. Our framework is written in Python and PyTorch. The source code is available online as an open-source project in Github.

\section{Related works and comparison}
\label{sec:related_works}

The field of model compression has grown enormously in the recent years, resulting in plethora of algorithmic approaches, research projects and software. In this section we limit our attention to the software aspect of the neural network compression. We discuss what kind of compression schemes are supported, available codes, and recently proposed compression frameworks.

\paragraph{Individual compressions}

The majority of neural network compression code is available as individual projects and recipes tailored for a particular compression and model. Usually it is released as a companion code for published research paper, e.g., codes of \cite{Courbar_15a,Tai_16a,Xu_18a,TungMori18a} and others.  Some repositories combine several compression recipes in a single place: e.g.,  Tensorpack\footnote{\url{https://github.com/tensorpack/tensorpack/tree/master/examples}} or the fork of the Caffe library by Wei Wen\footnote{\url{https://github.com/wenwei202/caffe}}. 

Out of many individual compressions proposed in the literature, the quantization aware training of \cite{Jacob_18a} has gained popularity and became a standard feature of major deep-learning frameworks. TensorFlow, Pytorch and MxNet natively support both training of such quantized models and allow an efficient inference afterwards. 

\paragraph{Efficient inference frameworks}

Relatively mature software is available if the goal is not to compress the model in a lossy way, but to run it unchanged as efficiently as possible on a given hardware. Such frameworks allow to convert (compile) already trained neural network to utilize the hardware-enabled fast computations: through usage of edge TPU-s on Pixel 4 (Pixel Neural Core) or Neural Engine on iPhone 8. Examples include Tensorflow Light\footnote{\url{https://www.tensorflow.org/lite}}, PyTorch Mobile\footnote{\url{https://pytorch.org/mobile/home/}}, Apple Core ML\footnote{\url{https://developer.apple.com/documentation/coreml}}, Qualcomm Neural Processing SDK\footnote{\url{https://developer.qualcomm.com/software/qualcomm-neural-processing-sdk}}, QNNPack\footnote{\url{https://engineering.fb.com/ml-applications/qnnpack/}} and others.

\paragraph{Compression frameworks}

The diversity compression mechanisms and limited support by deep learning frameworks led to development of specialized software libraries such as Distiller \cite{Zmora_19a} and NCCF \cite{Kozlov_20a}. Distiller and NCCF gather multiple compression schemes and corresponding training algorithms into a single framework, and make it easier to apply to new models. Both frameworks allow to apply multiple compression simultaneously to disjoint parts of single model. However, the underlying compression algorithms do not share same algorithmic base and might require deeper understanding from end user to efficiently tune the settings. 

\paragraph{What makes our approach special?}

Our approach is based on solid optimization principles, with guarantees of convergence under standard assumptions. It formulates the problem of model compression in a way that is intuitive and amenable to efficient optimization. The form of the actual algorithm is obtained systematically by judiciously applying mathematical transformations to the objective function and constraints. For example, if one wants to optimize the cross-entropy over a certain type of neural net, and represent its weights via a quantized codebook, then the L and C steps necessarily take a specific form. If one wants instead to represent the weights via low-rank matrices, a different C step results, and so on. The resulting algorithm is not based on combining backpropagation training with heuristics, such as pruning weights on the fly, which may result in suboptimal results or even non-convergence. The user does not need to work out the form of individual L or C steps (unless so desired), as we provide a range to choose from.

The LC algorithm is efficient in runtime; it does not take much longer than training the reference, uncompressed model in the first place. The compressed models perform very competitively and allow the user to easily explore the space of prediction accuracy of the model vs compression ratio (which can be defined in terms of memory, inference time, energy or other criteria). Our code has been extensively tested since 2017 through usage in internal research projects, and has resulted in multiple publications that improve the state of the art in several compression schemes \cite{CarreirIdelbay18a,IdelbayCarreir20a,IdelbayCarreir20c}.

But what truly makes the approach practical is its flexibility and extensibility. If one wants to compress a specific type of model with a specific compression scheme, all is needed is to pick a corresponding L step and C step. It is not necessary to create a specific algorithm to handle that choice of model and compression. Furthermore, one is not restricted to a single compression scheme; multiple compression schemes (say, low-rank plus pruning plus quantization) can be combined automatically, so they best cooperate to compress the model. The compression schemes that our code already supports make it possible for a user to mix and match them as desired with minimal effort. We expect to include further schemes in the future, as well as a range of model types.

\section{Model compression as a constrained optimization problem}

In this section, we briefly introduce the Learning-Compression \cite{Carreir17a} framework, which is the backbone of our software. Let us begin by assuming we have a previously trained model with weights $\w$, which were obtained by minimizing some loss function $L(\w)$. This is our \emph{reference} model, which represents the best loss we can achieve without compression. Here we omitted the exact definition of the weights $\w$, but for now, let us assume it has $P$ parameters. The ``Learning-Compression'' paper \cite{Carreir17a} defines the \emph{compression} as finding a low-di\-men\-sio\-nal parameterization $\bDelta(\bTheta)$ of the weights \w\ in terms of $Q$-sized parameter \bTheta\@, with $Q < P$.

We regard compression and decompression as mappings, while in the signal processing literature they are usually seen as algorithms (e.g., Lempel-Ziv algorithm \cite{ZivLempel77a}). Formally, the \emph{decompression mapping} $\bDelta$ maps a low-dimensional parameters \bTheta\@ to uncompressed model weights \w:
\begin{equation*}
	\bDelta\mathpunct{:}\ \bTheta \in \bbR^Q \rightarrow \w \in \bbR^P,
\end{equation*}
and the \emph{compression mapping} behaves as its ``inverse'':
\begin{equation*}
	\bPi(\w) = \smash{\argmin_{\bTheta}{\norm{\w - \bDelta(\bTheta)}^2}}.
\end{equation*}
 The goal of model compression is to find such \bTheta{} that its corresponding decompressed model has (locally) optimal loss. Therefore the \emph{model compression as a constrained optimization} problem is defined as: 
\begin{equation}
	\textcolor{black}{\min_{\w,\bTheta}{\, L(\w) } \quad \text{s.t.} \quad \w = \bDelta(\bTheta)}\label{eq:mcco}.
\end{equation}

The problem in eq.~\ref{eq:mcco} is constrained, nonlinear, and usually non-differentiable wrt \bTheta{} (e.g., when compression is binarization). To efficiently solve it, the LC-algorithm is obtained by converting this problem to an equivalent formulation using penalty methods (quadratic penalty or augmented Lagrangian) and employing an alternating optimization. This results in an algorithm that alternates two generic steps while slowly driving the penalty parameter $\mu \rightarrow \infty$:

\begin{itemize}
	\item \textbf{L (learning) step:} $\smash{\colbf{\min_{\w}{ L(\w) + \frac{\mu}{2} \norm{\w - \bDelta(\bTheta)}^2 }}}$. This is a regular training of the uncompressed model but with a quadratic regularization term. \emph{This step is independent of the compression type.}
	\item \textbf{C (compression) step:} $\smash{\textcolor{blue}{\min_{\bTheta}{ \norm{\w - \bDelta(\bTheta)}^2 } \Leftrightarrow \bTheta = \bPi(\w)}}$. This means finding the best (lossy) compression of $\w$ (the current uncompressed model) in the $\ell_2$ sense (orthogonal projection on the feasible set), and corresponds to our definition of the compression mapping \bPi. \emph{This step is independent of the loss, training set and task.} 
\end{itemize}

\begin{figure}[b!]
  \centering
    \psfrag{XX}[][]{~~\caja{c}{c}{$\overline{\w}$ \\ (reference)}}
    \psfrag{FF}[l][Bl]{\caja{c}{c}{$\w^*$ (optimal \\ compressed)}}
    \psfrag{Fd}[l][Bl]{\caja{c}{c}{$\bDelta(\bTheta^{\text{DC}})$ \\ (direct \\ compression)}}
    \psfrag{R}[lt][lt]{\hspace*{-2ex}\caja{t}{c}{\w-space \\ (uncompressed \\ models)}}
    \psfrag{w_mu}[r][r]{$\w^*(\mu)$}
    \psfrag{mappings}[][]{\caja{c}{c}{feasible models \calC \\ (decompressible \\ by \bDelta)}}
    \includegraphics[width=0.70\linewidth,bb=210 580 435 838,clip]{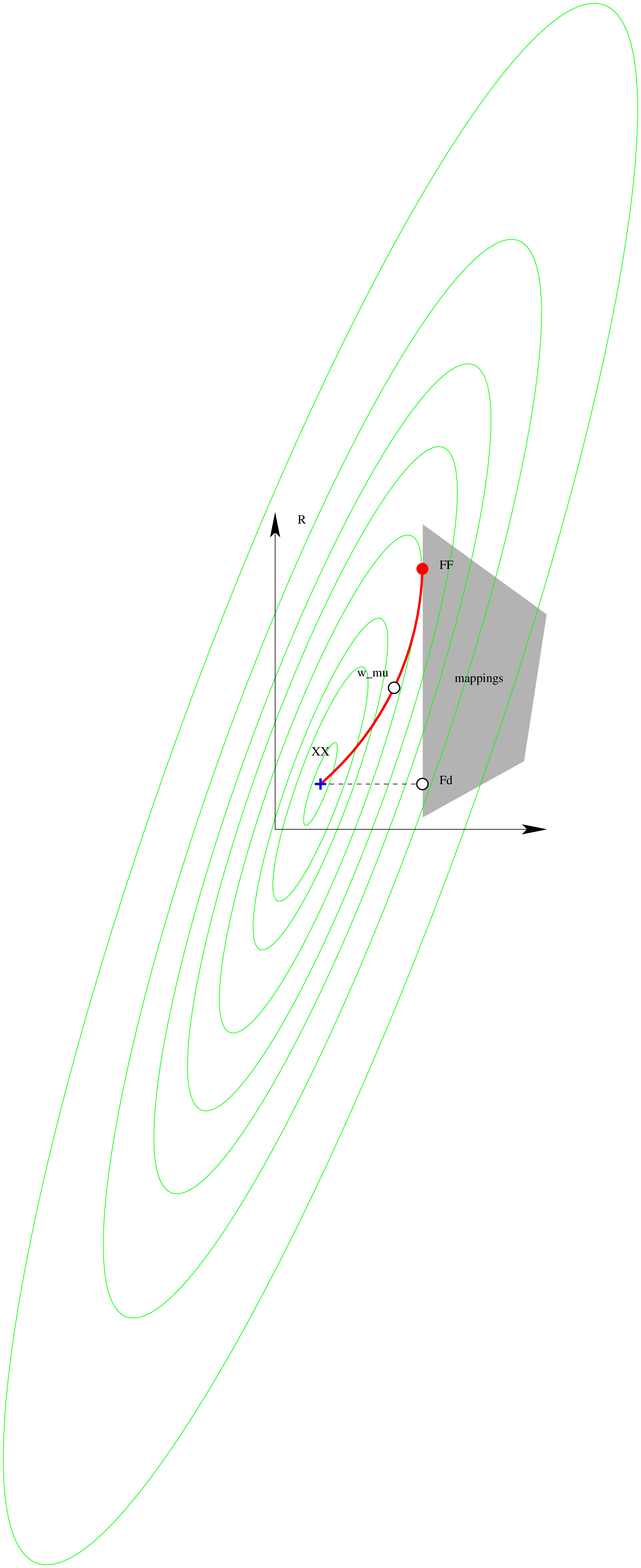}
  \caption{Schematic representation of the idea of model compression by constrained optimization. The plot illustrates the uncompressed model space (\w-space $= \bbR^P$), the contour lines of the loss $L(\w)$ (green lines), and the set of compressed models (the feasible set $\calC = \{\w \in \bbR^P\mathpunct{:}\ \w = \bDelta(\bTheta) \text{ for } \bTheta \in \bbR^Q\}$, grayed areas), for a generic compression technique \bDelta. The \bTheta-space is not shown. $\overline{\w}$ optimizes $L(\w)$ but is infeasible (no \bTheta\ can decompress into it). The direct compression $\w^{\text{DC}} = \bDelta(\bTheta^{\text{DC}})$ is feasible but not optimal compressed (not optimal in the feasible set). $\w^* = \bDelta(\bTheta^*)$ is optimal compressed. The red curve is the projection in \w-space of the solution path of the LC algorithm $(\w^*(\mu),\bTheta^*(\mu))$ for $\mu \ge 0$. See more details in \cite{Carreir17a}.}
  \label{f:LC-illustration}
\end{figure}

We will be using the quadratic penalty (QP) formulation throughout this paper to make derivations easier. In practice, we implement the augmented Lagrangian (AL) version which has additional vector of Lagrange multipliers $\blambda$, see Figure.~\ref{f:LC-pseudocode}. The QP version can be obtained from AL version by setting $\blambda=\0$ and skipping the multipliers update step. Fig.~\ref{f:LC-illustration} illustrates the idea of model compression as constrained optimization and the LC algorithm.

Our software capitalizes on the separation of the L and C steps: to apply a new compression mechanism under the LC formulation, the software requires only a new C step corresponding to this mechanism. Indeed, the compression parameter \bTheta\@ enters the L step problem as a constant regardless of the chosen compression type. Therefore, all L steps for any combination of compressions have the same form. Once the L step has been implemented for a model, any possible compression (C steps) can be applied.

More importantly, this separation allows using the best tools available for each of the L and C steps. For modern neural networks, the optimization of the L step means iterations over the dataset and requires solving it using SGD and hardware accelerators. The formulation of the C step, on the other hand, is given by $\ell_2$ minimization, and as we will see in the next chapter, solutions of it involve efficient algorithms. In fact, for certain compression choices, the C step problem is well studied and has a history of its usage on its own merit in the fields of data and signal compression. From the software engineering perspective, the separation of L and C steps makes code more robust and allows us to thoroughly test and debug each component separately.

\begin{figure}
	\vspace*{-3ex}
	\centering
	\small
	\hspace*{-1ex}
	\begin{tabular}{@{}c@{\hspace{0.02\linewidth}}|c@{}}
	
		\begin{minipage}[b]{0.48\linewidth}
			\vspace{0pt}
			\begin{tabbing}
			  n \= n \= n \= n \= n \= \kill
			\underline{\textbf{input}} training data and model with parameters \w \\
			$\colbf{\w} \leftarrow \overline{\w} = \argmin_{\w}{ L(\w) }$ \` {\small\textsf{pretrained model}} \\
			$\colit{\bTheta} \leftarrow \bTheta^{\text{DC}} = \bPi(\overline{\w})$ \` {\small\textsf{init compression}} \\
			$\colemph{\blambda} \leftarrow \0$ \\
			\underline{\textbf{for}} $\mu = \mu_0 < \mu_1 < \dots < \infty$ \+ \\
			\colbf{$\w \leftarrow \argmin_{\w}{ L(\w) + \smash{\frac{\mu}{2} \norm{\smash{\w - \bDelta(\bTheta) - \frac{1}{\mu} \blambda}}^2} }$} \` \colbf{{\textsf{L step}}} \\
			\colit{$\bTheta \leftarrow \argmin_{\bTheta}{ \norm{\w - \smash{\frac{1}{\mu} \blambda} - \bDelta(\bTheta)}^2 }$} \` \colit{{\textsf{C step}}} \\
			\colemph{$\blambda \leftarrow \blambda - \mu (\w - \bDelta(\bTheta))$} \` \colemph{{\textsf{multipliers step}}} \\
			\textbf{if} $\norm{\w - \bDelta(\bTheta)}$ is small enough \textbf{then} exit the loop \- \\
			\underline{\textbf{return}} \w, \bTheta
		\end{tabbing}
		\end{minipage}
		&
		\begin{minipage}[b]{0.5\linewidth}{
			\begin{lstlisting}[numbers=none, gobble=6, tabsize=2]
			class LCAlgorithm():
				# Housekeeping code is skipped
				# Pretrained model is provided by user at init
				def run(self):
					self.mu = 0						
					self.c_step(step_number=0)	

					for step_n, mu in enumerate(self.mu_schedule):
						self.mu = mu
						self.l_step(step_n) # call user-provided L step
						self.c_step(step_n) # resolve compression tasks
						self.multipliers_step() 
			\end{lstlisting}
		\vspace{-2em}}
		\end{minipage}
	\end{tabular}
	\caption{\emph{Left}: pseudocode of the LC algorithm using the augmented Lagrangian. \emph{Right}: corresponding implementation in the \texttt{LC\-Algorithm} class, the main running method is shown.}
	\label{f:LC-pseudocode}
\end{figure}

\section{Supported compressions}
\label{sec:supported_compression}

In this section, we describe some of the compression schemes supported by our library. The complete list of supported compressions is given in Table~\ref{tab:supported_compressions}. We expect to add more compressions in the future.

\subsection{Quantization}

The quantization is the process of reducing the precision of the weights, and achieved by imposing a constraint on each weight $w_i$ to belong to set of values $\calC$ --- the codebook. Depending on the allowed values in the codebook, the quantization schemes are known under different names: binarization --- when $\calC =\{0,1\}$ or $\{-1,1\}$, ternarization with $\calC=\{-1,0,+1\}$, the powers-of-two scheme with $\calC =\{ 0, \pm 1, \pm 2, \dots, \pm 2^{s-1}\}$, and others. 

\par
Let us consider the general case when we compress the weights of the model with a learned codebook of size $K$, i.e., $\calC = \{c_1, c_2, \dots, c_K\}$. We will use the equivalent formulation of the quantization using a binary assignment variable $\z_i$ ($\sum_k z_{ik} = 1$) for each weight $w_i$. Then our compression goal is:
\begin{equation*}
	\min_{\w, \calC, \z_1, \dots \z_P} L(\w) \quad \text{s.t.} \quad w_i = \sum_{k=1}^K z_{ik} c_k, \quad  \forall i=1\dots P.
\end{equation*}
This formulation immediately falls into the Learning-Compression form of eq.~\ref{eq:mcco} with $\bTheta=(\calC, \z_1, \dots \z_P)$. The corresponding C step problem of $\min_{\bTheta} \norm{\w-\bDelta(\bTheta)}^2 $ has the form of:
\begin{equation}
\label{eq:k_means}
	\min_{\calC,\, \z_1, \dots \z_P} \sum_{i=1}^P \sum_{k=1}^{K} z_{ik} (w_i-c_k)^2,
\end{equation}
which has been thoroughly studied in signal compression and unsupervised clustering literature, and known as the $k$-means clustering problem. The general $k$-means problem is NP hard \cite{DasgupFreund09a, Aloise_09a}, however, this is a scalar version which has an efficient globally optimal solution using dynamic programming \cite{Bruce65a,WuRokne89a,Wu91a}. 

Our software provides both $k$-means and dynamic programming solutions for the C step of adaptive quantization problem (eq.~\ref{eq:k_means}). Additionally, we provide solutions for some fixed- and scaled- binarization and ternarization schemes as listed in Table~\ref{tab:supported_compressions} (see \cite{CarreirIdelbay17a} for exact details).

\begin{figure}[t]
  \vspace*{-2ex}
  \centering
  \psfrag{P}[][]{codebook size, $K$}
  \psfrag{14}{14}
  \psfrag{12}{12}
  \psfrag{10}{10}
  \psfrag{2}{2}
  \psfrag{4}{4}
  \psfrag{8}{8}
  \psfrag{16}{16}
  \psfrag{32}{32}
  \begin{tabular}{@{}c@{\hspace*{1em}}c@{}}
    \hspace*{1em} Quantization of VGG16 & \hspace*{1em} Weight pruning of ResNets \\
    \psfrag{E}[B][]{test error $E_{\text{test}}$ (\%)}
    \psfrag{learned}[l][l]{{LC}}
    \psfrag{direct}[l][l]{{quant.$\rightarrow$ retrain}}
    \psfrag{reference}[l][l]{{reference}}
    \psfrag{R}[l][l]{{\textbf{R}}}
    \includegraphics*[width=0.50\linewidth,clip]{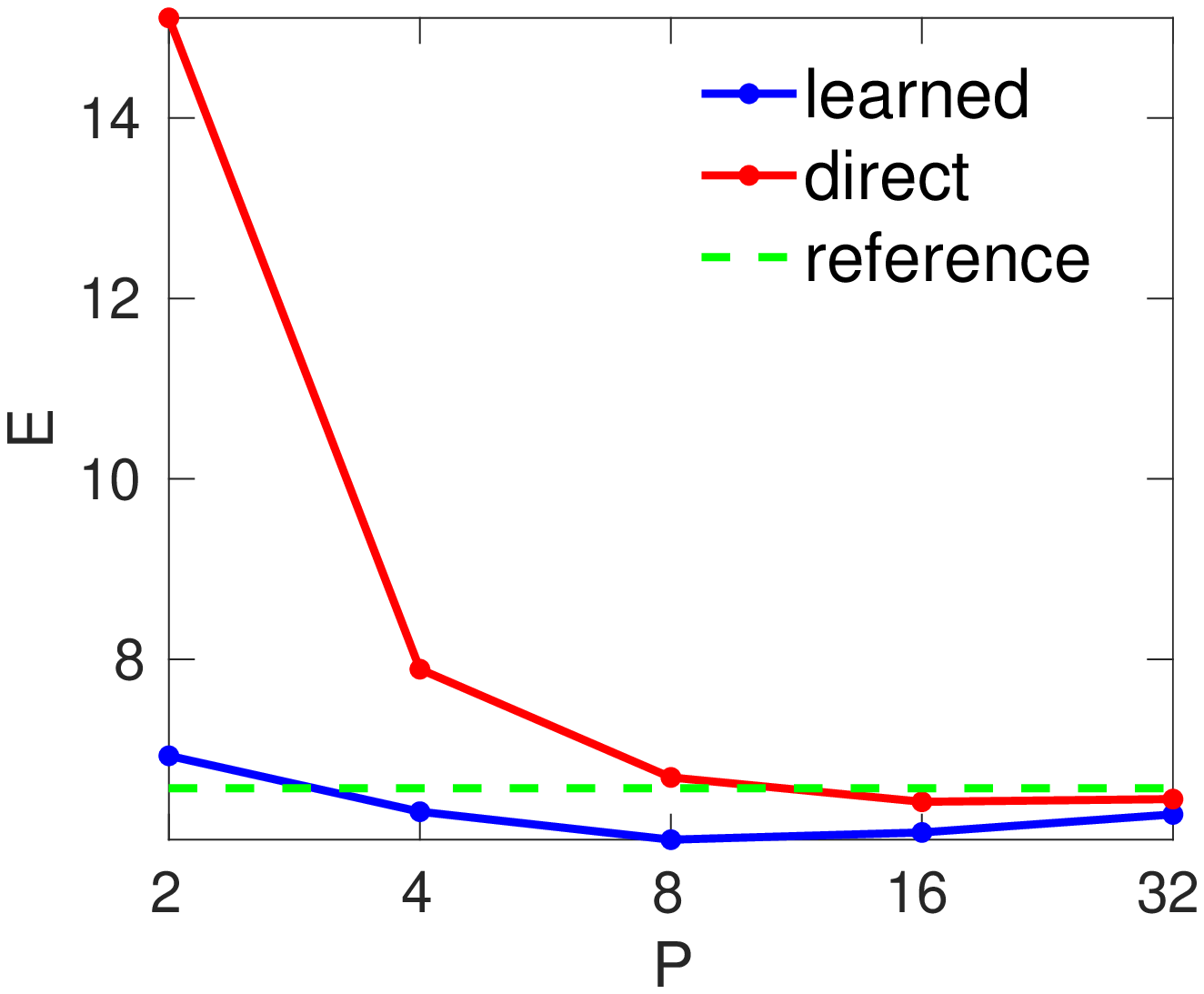} & \psfrag{E}[B][]{test error $E_{\text{test}}$ (\%)}
  \psfrag{P}[][]{proportion of surviving weights $P$\%}
  \psfrag{resnet32}[l][l]{\large ResNet32}
  \psfrag{resnet56}[l][l]{\large ResNet56}
  \psfrag{resnet110}[l][l]{\large ResNet110}
  \psfrag{14}{14}
  \psfrag{12}{12}
  \psfrag{10}{10}
  \psfrag{8}{8}
  \psfrag{5}{5}
  \psfrag{15}{15}
  \includegraphics*[width=0.5\linewidth, clip]{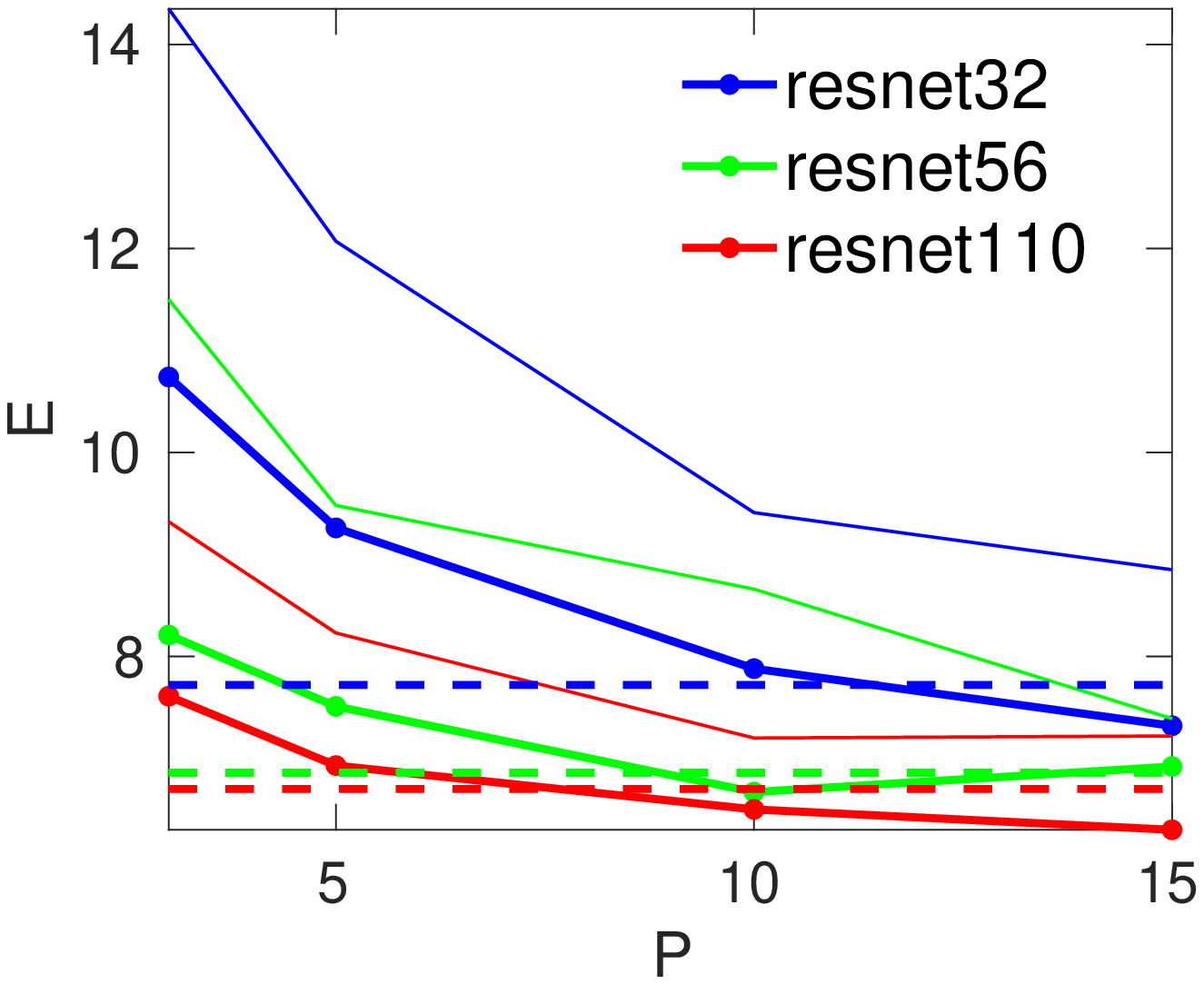}
  \end{tabular}
  \vspace*{-2ex}
  \caption{\emph{Left:} Tradeoff of quantizing VGG16 trained on CIFAR10. Results of LC is given by blue curve, and compared to quantize$\rightarrow$retrain approach similar to \cite{Han_16a} (red curve). Details of experiments are in \cite{CarreirIdelbay17a}. \emph{Right:} Weight pruning of ResNets trained on CIFAR10 dataset using $\ell_0$-constraint formulation of LC (thick lines) and comparison to magnitude based pruning with retraining (thin lines). Horizontal dashed lines correspond to reference test errors of respective networks. Full details of experiments can be found in \cite{CarreirIdelbay18a}.}
  \label{f:VGG_ResNet_tradeoff}
  \vspace*{-1ex}
\end{figure}

\subsection{Pruning}

Pruning is the process of removing (or sparsifying) the weights of the model. One way of formulating this problem is by using the sparsification penalties and constraints, e.g., $\ell_0$ or $\ell_1$, which will limit the number of allowed non-zero weights. A particularly useful pruning scheme is $\ell_0$-norm ($\norm{\cdot}_0$) constrained pruning defined as:
\begin{equation}
	\label{eq:l0_purning}
	\min_{\w} \, L(\w) \quad \text{s.t.} \quad \norm{\w}_0 \leq \kappa.
\end{equation}
Since the $\ell_0$-norm measures the number of non zero items in the vector, the formulation of eq.~\ref{eq:l0_purning} allows to precisely specify the number of remaining weights.

To bring it into the Learning-Compression form (eq.~\ref{eq:mcco}) we introduce a copy parameter $\btheta$ and obtain an equivalent optimization problem:
\begin{equation*}
	\min_{\w} \, L(\w) \quad \text{s.t.} \quad \w=\btheta, \quad \norm{\btheta}_0 \leq \kappa.
\end{equation*}
for which the C step is given by solving:
\begin{equation}
	\label{eq:l0_purning_cstep}
	\min_{\theta} \norm{\w-\btheta}^2 \quad \text{s.t.} \quad \norm{\btheta}_0 \leq \kappa .
\end{equation}
The solution of eq.~\ref{eq:l0_purning_cstep} can be obtained by selecting all but top-$\kappa$ weights (in magnitude) of $\w$ and zeroing remaining. 

Using similar steps, we can obtain the C steps for $\ell_1$ constrained formulation of pruning, and extend it to penalty based forms, e.g. $\min_{\w} L(\w) + \lambda \norm{\w}_0$, see \cite{CarreirIdelbay18a} for further details. In our framework we provide the implementation for all combinations of $\ell_0$ and $\ell_1$-norms, both constraint and penalty.

\subsection{Low-rank compression}

Our framework supports compressing the weight matrices of each layer to a given (preselected) target rank. This allows parametrizing the resulting compressed weight matrix $\W$ as a product of $\U\V^T$. However, such compression requires knowing the right choice of the ranks, or otherwise, it will affect the error-compression tradeoff of the resulting model. To alleviate this issue, we include the implementation of the automatic rank selection from \cite{IdelbayCarreir20a}, which we describe next.

Assume we have a reference model with $L$ layers and the weights $\w=\{\W_1, \dots, \W_L \}$, where $\W_l$ is the weight matrix of layer $l$. We want to optimize the following model selection problem over possible low-rank models:
\begin{equation*}
	\min_{\w} L(\w) + \lambda \, C(\w) \quad \text{s.t.} \quad \rank{\W_l} = r_l \leq R_l, \quad \forall \,l=1\dots L
\end{equation*}
here $R_l$ is the maximum possible rank for matrix $\W_l$. The compression cost $C(\w)$ is defined in terms of the ranks of the matrices:
\begin{equation*}
  C(\w) = \alpha_1 C(r_1) + \alpha_2 C(r_2) + \dots + \alpha_L C(r_L),
\end{equation*}
can capture both storage bits to save space, or total floating point operations to speed up the model. To put it into Learning-Compression form (eq.~\ref{eq:mcco}), we introduce the parameter $\bTheta_l$ for each layer, with constraint $\W_l = \bTheta_l$. Then, the objective of the C step separates over layers into:
\begin{equation*}
	\begin{split}
		\label{e:Cstep1}
		\min_{\bTheta_l,r_l} \ \ & {\lambda \, C_l(r_l) + \frac{\mu}{2} \norm{\W_l - \bTheta_l}^2 } \\
		\text{s.t.} \ \ & \rank{\bTheta_l} = r_l \le R_l.
	\end{split}
\end{equation*}
The solution of this C step was given in \cite{CarreirIdelbay18a}, and involves an SVD and enumeration over the ranks for each layer's weight matrix.

\begin{figure}[t!]
  \centering
  \psfrag{L}[B][]{training loss $L$}
  \psfrag{err}[B][]{test error $E_{\text{test}}$ (\%)}
  \psfrag{flops}[][t]{MFLOPs}
  \psfrag{resnet20}[l][l]{\Large ResNet20}
  \psfrag{ResNet20}[l][l]{\Large ResNet20}
  \psfrag{resnet32}[l][l]{\Large ResNet32}
  \psfrag{ResNet32}[l][l]{\Large ResNet32}
  \psfrag{resnet56}[l][l]{\Large ResNet56}
  \psfrag{ResNet56}[l][l]{\Large ResNet56}
  \psfrag{resnet110}[l][l]{\Large ResNet110}
  \psfrag{ResNet110}[l][l]{\Large ResNet110}
  \psfrag{vgg16}[l][l]{\Large VGG16}
  \psfrag{VGG16}[r][r]{\Large VGG16}
  \psfrag{nin}[l][l]{\Large NIN}
  \psfrag{NIN}[l][l]{\Large NIN}
  \psfrag{Wen17}[cc][lc]{\tiny 1}
  \psfrag{Ye18-b}[cc][lc]{\tiny 2}
  \psfrag{Ye18-a}[cc][lc]{\tiny 2}
  \psfrag{Zhuang18-a}[cc][lc]{\tiny 3}
  \psfrag{Zhuang18-b}[cc][lc]{\tiny 3}
  \psfrag{Li17}[cc][lc]{\tiny 4}
  \psfrag{Li17a}[cc][lc]{\tiny 4}
  \psfrag{Li17b}[cc][lc]{\tiny 4}
  \psfrag{He17}[cc][lc]{\tiny 5}
  \psfrag{Yu18}[cc][lc]{\tiny 6}
  \psfrag{Xu18}[cc][lc]{\tiny 7}  
  \psfrag{cobla}[cc][lc]{\tiny 8} 
  \psfrag{1a}[lc][cc]{1 --- Wen et~al.~\cite{Wen_17a}}  
  \psfrag{2a}[lc][cc]{2 --- Ye et~al.~\cite{Ye_18a}}
  \psfrag{3a}[lc][cc]{3 --- Zhuang et~al.~\cite{Zhuang_18a}}
  \psfrag{4a}[lc][cc]{4 --- Li et~al.~\cite{Li_17b}}
  \psfrag{5a}[lc][cc]{5 --- He et~al.~\cite{He_17a}}
  \psfrag{6a}[lc][cc]{6 --- Yu et~al.~\cite{Yu_18a}}
  \psfrag{7a}[lc][cc]{7 --- Xu et~al.~\cite{Xu_18a}}
  \psfrag{8a}[lc][cc]{8 --- Li and Shi~\cite{LiShi18a}}
  \psfrag{R}[cc][lc]{{\textbf{R}}}
  \includegraphics[width=1\linewidth,clip]{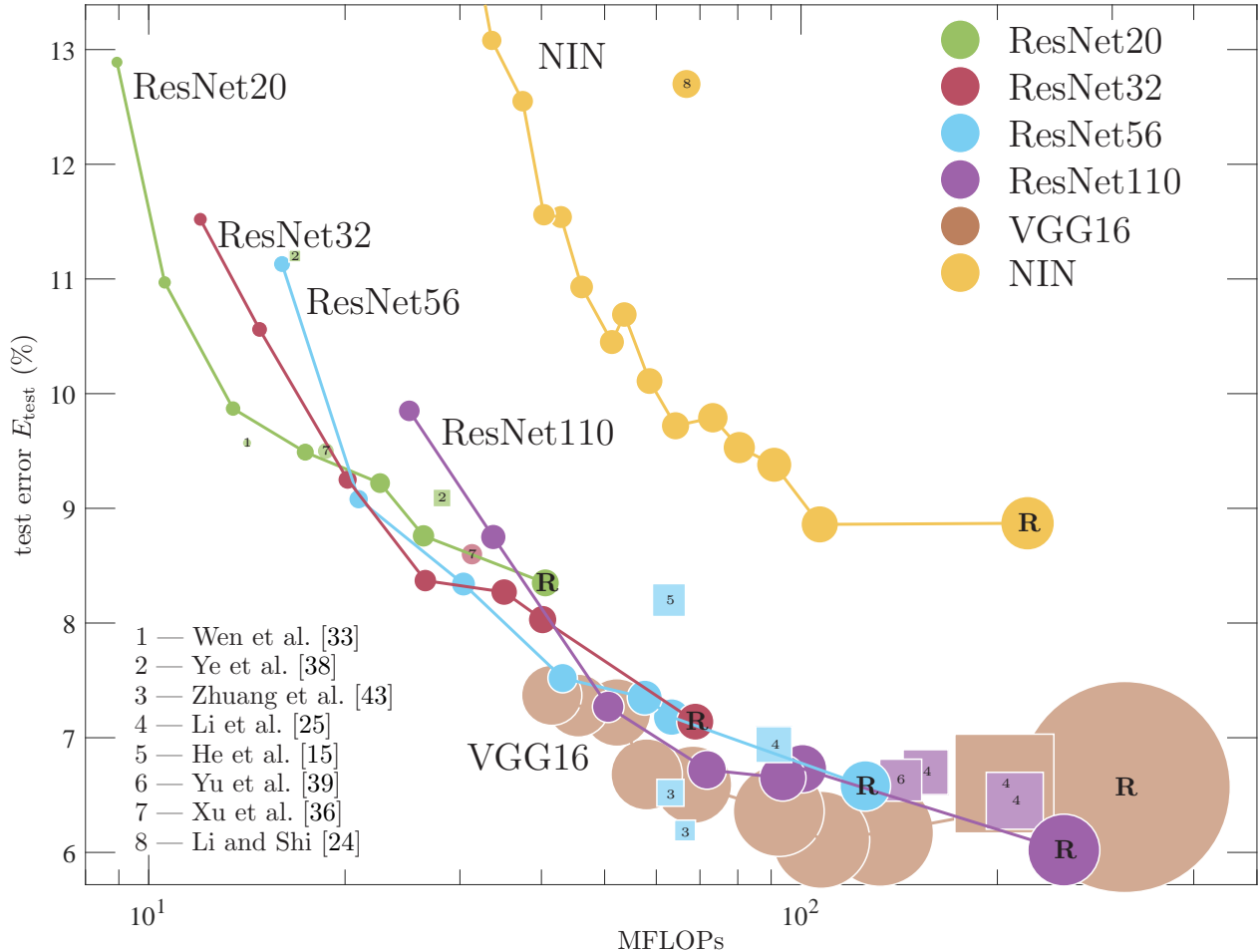}
  \caption{Error-compression space of test error (Y axis), inference FLOPs (X axis) and number of parameters (ball size for each net), for multiple networks trained on CIFAR10 and compressed with low-rank and structured pruning methods. Results of rank selection with LC algorithm over different $\lambda$ values for a given network span a curve, shown as connected circles \mbox{\raisebox{-0.8ex}[0pt][0pt]{{\Huge$\bullet$}}\hspace*{-0.35ex}---\hspace*{-0.35ex}\raisebox{-0.8ex}[0pt][0pt]{{\Huge$\bullet$}}}, which starts on the lower right at the reference \textbf{R} ($\lambda = 0$) and then moves left and up. Other published results using low-rank compression are shown as isolated circles labeled with a citation. Other published results involving structured filter pruning are shown as isolated squares labeled with a citation. The area of a circle or square is proportional to the number of parameters in the corresponding compressed model. Ideal models are small balls (having few parameters) on the left-bottom (where both error and FLOPs are the smallest). See \cite{IdelbayCarreir20a} for full details of experiments.}
  \label{f:all_tradeoff}
\end{figure}

\begin{table}[t]
	\center
	\begin{tabular}{@{}ll@{}}
		\toprule
		Type & Forms \\
		\midrule
		\multirow{3}{3cm}{Quantization} & Adaptive Quantization into $\{c_1, c_2, \dots c_K\}$ \\
										& Binarization into $\{-1, 1\}$ and $\{-c, c\}$ \\
										& Ternarization into $\{-c, 0, c\}$ \\
										\cmidrule{2-2}
		\multirow{4}{3cm}{Pruning} 		& $\ell_0$-constraint (s.t., $\norm{\w}_0 \leq \kappa$) \\
										& $\ell_1$-constraint (s.t., $\norm{\w}_0 \leq \kappa$) \\
										& $\ell_0$-penalty ($\alpha \norm{\w}_0 $) \\
										& $\ell_1$-penalty ($\alpha \norm{\w}_1 $) \\
										\cmidrule{2-2}
		\multirow{3}{3cm}{Low-rank}		& Low-rank compression to a given rank \\
										& Low-rank with \emph{automatic} rank selection for FLOPs reduction \\
										& Low-rank with \emph{automatic} rank selection for storage compression \\
										\cmidrule{2-2}
		\multirow{4}{4cm}{Additive Combinations} & Quantization + Pruning \\
										& Quantization + Low-rank \\
										& Pruning + Low-rank \\
										& Quantization + Pruning + Low-rank\\
		\bottomrule
	\end{tabular}
	\caption{Currently supported compression types, with their exact forms. These compression can be defined per one or multiple layers, and different compression can be applied to different parts of the model.}
	\label{tab:supported_compressions}
\end{table}

\section{Design of the software}

Equipped with Learning-Compression algorithm and some building-block compressions, we now discuss the design of our library.
The main goals are to have an easy to use, efficient, robust, and configurable neural network compression software. Particularly, we would like to have a flexibility of applying any available compression (Table \ref{tab:supported_compressions}) to any parts of the neural network with per-layer granularity:
\begin{itemize}[itemsep=-0.4em]
	\item a single compression per layer (e.g., low-rank compression for layer 1 with target rank 5)
	\item a single compression per multiple layers (e.g. prune 5\% of weights in layer 1 and 3, jointly)
	\item mixing multiple compressions (e.g., quantize layer 1 and prune jointly layers 2 and 3)
	\item additive compressions
\end{itemize}
To implement such desiderata, we leverage the modularity of the LC algorithm and introduce some additional building blocks in between.

\paragraph{L step} 
We hand off the model training operations, the L step, to the user through the lambda functions. This gives a fine-grained control on the model's actual learning, utilization of hardware, pulling the data sources, and other necessary steps required for training. Usually, the implementation of the L step is already available or can be extracted from the training code used for the reference (uncompressed) model. On the left of Figure~\ref{fig:typical_l_and_c_steps} we give a typical way of implementing the L step in PyTorch.

\begin{figure}[b!]
\begin{tabular}{c|c}
\begin{lstlisting}[language=Python, 
	xleftmargin=.2\textwidth]
def my_l_step(model, lc_penalty, args**):
    # ... skipped ...
    loss = model.loss(out_, target_) + lc_penalty()
    loss.backward()
    optimizer.step()
    # ... skipped ...
\end{lstlisting}  \vspace{-0.5em}&

\begin{lstlisting}[language=Python, 
	xleftmargin=.2\textwidth
]
class ScaledBinaryQuantization(CompressionTypeBase):
    # Housekeeping code is skipped
    def compress(self, data):
        a = np.mean(np.abs(data))
        quantized = 2 * a * (data > 0) - a
        return quantized
\end{lstlisting} \vspace{-0.5em}
\end{tabular}
\caption{\emph{Left:} a typical implementation of the L step in PyTorch; some code (the optimizer and data source configurations) is skipped for brevity. \emph{Right:} the C step implementation. To add a new compression to the framework\, one needs to inherit from {\lstinline[basicstyle=\ttfamily]+CompressionTypeBase+} class and implement {\lstinline[basicstyle=\ttfamily]+compress+} function.}
\label{fig:typical_l_and_c_steps}
\end{figure}

\paragraph{C step} 

All provided compressions of Table \ref{tab:supported_compressions} are implemented as subclasses of \texttt{CompressionTypeBase} class, and the actual C step is exposed through the \texttt{compress} method. This allows a straightforward extension of the library of compressions: if needed, the user simply wraps the custom C-step solution into an object of \texttt{CompressionTypeBase} class. For example, on the right of Figure~\ref{fig:typical_l_and_c_steps} we show how a new quantization can be implemented.

\paragraph{Compression tasks}

To instruct the framework on which compression types should be applied to which parts of the model, the user needs to populate a compression tasks structure. This structure is a list of simple mappings of the form: (parameters) $\rightarrow$ (compression view, compression type), which is implemented as a python dictionary. The \emph{parameters} are any subset of model weights, which are wrapped into internal \texttt{Parameter} object. The \emph{compression view} is another internal structure that handles reshaping of the model weights into a form suitable for compression, e.g., reshaping the weight tensor of a convolutional layer into a matrix for low-rank compression.

While our strategy of defining the compression tasks might seem unnecessarily complicated, it brings \emph{a considerable amount of flexibility}. For instance, it erases the limitations of standard compression approaches with coarse layer-based granularity: we can compress multiple layers with a single compression, or a single layer with multiple compressions, while simultaneously mixing different compressions in a single model. This abstraction disentangles compression from the model structure and allows us to construct complicated schemes of compressions in a \emph{mix-and-match} way.  For example, user can jointly compress a three-layer neural network so that the first and third layers are quantized with the same codebook, and the second layer is a low-rank matrix simultaneously, see Figure~\ref{fig:compression_tasks_semantics} and other examples in section~\ref{sec:showcase}. Such fine-grained control allows to include expert knowledge about properties of a particular model (e.g., do not quantize the first layer) without much effort.

\begin{figure}[b]
\centering
\begin{tabular}{l  l}
\multirow{2}{*}{Semantics:} & (layer 1, layer 3) $\rightarrow$ (as a vector, adaptive quantization $k=6$), \\
& (layer 2) $\rightarrow$ (as is, low-rank with $r=3$) \\[0.5em]
Python code: & 
{
\begin{lstlisting}[language=Python,numbers=none]
from lc.torch import ParameterTorch as Param, AsVector, AsIs
compression_tasks = {
    Param([l1.weight, l3.weight]): (AsVector, AdaptiveQuantization(k=6)),
    Param(l2.weight):              (AsIs,     LowRank(target_rank=3))
}
\end{lstlisting}\vspace*{-2em}
}
\end{tabular}
\caption{Semantics and the actual python code for compression tasks to quantize first and second layers of a NN with a single adaptive codebook of size $k=6$ and the third layer with a low-rank matrix of rank 3. Notice how the semantics translates almost verbatim into the python code.}
\label{fig:compression_tasks_semantics}
\end{figure}

\paragraph{Running the software}

To compress a model, the user needs to construct an \texttt{lc.Algorithm} object and provide:
\begin{enumerate}[itemsep=-0.2em]
\item a model to be compressed
\item associated compression tasks
\item implementation of the L step
\item a schedule of $\mu$ values, and
\item an evaluation function to keep track of the loss/error during the compression.
\end{enumerate}

\begin{lstlisting}[caption=Running the LC algorithm., xleftmargin=.2\textwidth, label={listing:run_lc}]
lc_alg = lc.Algorithm(
    model=net,                            # a model to compress
    compression_tasks=compression_tasks,  # specifications of compression
    l_step_optimization=my_l_step,        # implementation of the L step
    mu_schedule=mu_s,                     # schedule of the mu values
    evaluation_func=train_test_acc_eval_f # the evaluation function
)
lc_alg.run()                              # an entry point to the LC algorithm
\end{lstlisting}

Once the \texttt{run} method is called, the LC algorithm will start execution, at which point the library will proceed in line-by-line correspondence to the pseudocode on the left of Figure~\ref{f:LC-pseudocode}). Currently, each of the compression tasks (and corresponding C step) is called in order. Yet, due to the nature of the LC algorithm, every compression task's C steps can be run in parallel, further improving the efficiency of the algorithm.

\section{Showcase}
\label{sec:showcase}
In this section, we demonstrate the flexibility of our framework by easily exploring multiple compression schemes with minimal effort. As an example, say we are tasked with compressing the storage bits of the LeNet300 neural network trained on MNIST dataset (10 classes, $28\times 28$ gray-scale images). The LeNet300 is a three-layer neural network with 300, 100, and 10 neurons respectively on every layer; the reference has an error of 2.13\% on the test set.

In order to run the LC algorithm, we need to provide an L step implementation and compression tasks to an instance of the \texttt{LCAlgorithm} class as described in Listing~\ref{listing:run_lc}. We implement the L step below:

\begin{lstlisting}[language=Python,caption=Complete implementation of the L step for LeNet300 using Pytorch. The regular model training is exactly as the L step code with only difference of loss computation: the L step needs loss plus penalty, xleftmargin=.2\textwidth, label={listing:full_l_step}]
def my_l_step(model, lc_penalty, step):
    params = list(filter(lambda p: p.requires_grad, model.parameters()))
    lr = lr_base*(0.98**step) # we use a fixed learning rate for each L step
    optimizer = optim.SGD(params, lr=lr, momentum=0.9, nesterov=True)
    for epoch in range(epochs_per_step):
        for x, target in train_loader: # loop over the dataset
            optimizer.zero_grad()
            loss = model.loss(model(x), target) + lc_penalty() # loss + LC penalty
            loss.backward()
            optimizer.step()
\end{lstlisting}

\begin{table}[p]
  \centering
	\begin{tabular}{@{}m{3cm}>{\centering}m{10cm}c@{}}
	\toprule
	\multicolumn{1}{c}{Compression} & Code for compression tasks & \multicolumn{1}{c}{Error} \\
	\midrule
	no compression & & \begin{tabular}{@{}ll@{}}
	Train & 0.00\%\\
	Test  & 2.13\%\\
	\end{tabular} \\
	\midrule 
	quantize all layers & {
	\begin{lstlisting}[belowskip=-25pt]
	compression_tasks = {
		Param(l1.weight): (AsVector, AdaptiveQuantization(k=2)),
		Param(l2.weight): (AsVector, AdaptiveQuantization(k=2)),
		Param(l3.weight): (AsVector, AdaptiveQuantization(k=2))
	}
	\end{lstlisting}  \vspace{-1em}} & \begin{tabular}{@{}ll@{}}
	Train & 0.02\%\\
	Test  & 2.56\%\\
	\end{tabular} \\
	\midrule
	quantize first and third layers &
	\begin{lstlisting}[belowskip=-35pt]	
	compression_tasks = {
		Param(l1.weight): (AsVector, AdaptiveQuantization(k=2)),
		Param(l3.weight): (AsVector, AdaptiveQuantization(k=2))
	}
	\end{lstlisting}   & \begin{tabular}{@{}ll@{}}
	Train & 0.00\%\\
	Test  & 2.26\%\\
	\end{tabular} \\\\
	\midrule
	prune all but 5\% & 
	\begin{lstlisting}[belowskip=-30pt]
	compression_tasks = {
		Param([l1.weight, l2.weight, l3.weights]): 
			(AsVector, ConstraintL0Pruning(kappa=13310)) # 13310 = 5%
	}
	\end{lstlisting} & \begin{tabular}{@{}ll@{}}
	Train & 0.00\%\\
	Test  & 2.18\%\\
	\end{tabular} \\
	\midrule
	single codebook quantization with additive pruning of all but 1\% & 
	\begin{lstlisting}[belowskip=-30pt]
	compression_tasks = {
		Param([l1.weight, l2.weight, l3.weights]): [
			(AsVector, ConstraintL0Pruning(kappa=2662)), # 2662 = 1%
			(AsVector, AdaptiveQuantization(k=2))
		]
	}
	\end{lstlisting} & \begin{tabular}{@{}ll@{}}
	Train & 0.00\%\\
	Test  & 2.17\%\\
	\end{tabular} \\
	\midrule
	prune first layer, low-rank to second, quantize third  &
	\begin{lstlisting}[belowskip=-30pt]
	compression_tasks = {
		Param(l1.weight): (AsVector, ConstraintL0Pruning(kappa=5000)),
		Param(l2.weight): (AsIs,     LowRank(target_rank=10))
		Param(l3.weight): (AsVector, AdaptiveQuantization(k=2))
	}
	\end{lstlisting} &  \begin{tabular}{@{}ll@{}}
	Train & 0.04\%\\
	Test  & 2.51\%\\
	\end{tabular} \\
	\midrule
	rank selection with $\alpha=10^{-6}$  &
	\begin{lstlisting}[belowskip=-30pt]
	compression_tasks = {
		Param(l1.weight): (AsIs,     RankSelection(alpha=1e-6))
		Param(l2.weight): (AsIs,     RankSelection(alpha=1e-6))
		Param(l3.weight): (AsIs,     RankSelection(alpha=1e-6))
	}
	\end{lstlisting} 
	&  \begin{tabular}{@{}ll@{}}
	Train & 0.00\%\\
	Test  & 1.90\%\\
	\end{tabular} \\
	\bottomrule
	\end{tabular}
\caption{We are showcasing some of the possible compression tasks on the LeNet300 neural network trained on the MNIST dataset. Notice that trying a new combination of compressions is as simple as writing a new \emph{compression tasks} structure with a possible change of learning rates and $\mu$-schedule.}
\label{tab:showcase_lenet300}
\end{table}

Now, having the L step implementation, we can formulate the compression tasks. Say, we would like to know what would be the test error if the model is optimally quantized with a separate codebook on each layer? Test error in such case will be 2.56\%, which is 0.43\% higher than the reference. What would be the performance of the model if one would quantize only the first and the third layers, leaving the second layer untouched? Well, test error in such case will be 2.18\%. What about if we prune all but 5\% of the weights? Yes, the LC algorithm and our framework can handle all of these combinations and more; see Table~\ref{tab:showcase_lenet300} for details. We can even apply different compressions to every layer, e.g., take a look at the last row of Table~\ref{tab:showcase_lenet300}, where we apply quantization, pruning, and low-rank compression to the different parts of the LeNet300. This is possible with the only change required: we need to provide a new compression task for desired compression and possibly a new schedule of $\mu$-s and learning rates (if using SGD). Our framework allows to compare different compression techniques in single library with common algorithm.

In the compressions described in Table~\ref{tab:showcase_lenet300} we used exponential schedule of $\mu$ values: for all of quantization/pruning experiments the schedule was $\mu_i = 9\times10^{-5}\times 1.1^{i}$ at $i$-th L step and for those involving low-rank the schedule was $\mu_i = 9\times10^{-5}\times 1.4^{i}$. The learning rate for SGD decayed by 0.98 after every L step: for pure quantization experiments the base learning rate was 0.09, for pruning 0.1, and for mixture of schemes the learning rate was 0.05. We used total of 40 L steps, and every L step was running for 20 epochs.

\section{Practical advice}

We implemented the LC algorithm originally in 2017, and we have gone through multiple refinements and code reimplementations. We have applied it to compressing a wide array of relatively large neural nets, such as LeNet, AlexNet, VGG, ResNet, etc., which are themselves tricky to train well in the first place. In the process, we have gathered a considerable amount of practical knowledge on the behavior of the LC algorithm on both small and large models and datasets. We would like to share a list of common pitfalls so future users of our framework would hopefully avoid them:
\begin{description}
\item[Monitor the progression] of the algorithm. Specifically, two important quantities to keep an eye on:
  \begin{itemize}
  \item The loss of the L step: $L(\w) + \frac{\mu}{2}\norm{\w - \bDelta(\bTheta)}^2$. The total loss at the end of the L step must be smaller than the total loss at the beginning. Preferably, it must be much smaller. If some L step has not reduced the loss, optimization parameters of the step should be tuned.
  \item The loss of the C step, $\norm{\w - \bDelta(\bTheta)}^2$, must have a smaller value after each C step. This often fails when new compression is introduced into the pipeline, where \texttt{compress} method is not fully tested. For the base compressions in the framework, we made sure they always optimize the quadratic distortion loss of the C step.
  \end{itemize}
\item[L steps are not equally created] L steps can, and some of them should be optimized better than others. Here, by ``better'' we mean the reduction in the loss value should be evident. This advice is usually applicable for the first L steps. In case of using SGD, it is often helpful to train the first L step for a larger number of iterations than other steps.
\item[On learning rates of SGD] In the case of L step optimization happening by means of SGD, often good initial learning rates can be chosen by inspecting the learning rates used in the training of the original (uncompressed) reference model.
\item[On $\mu$ schedule] Theoretically, the sequence of $\mu$ values should start at $0$ and infinitesimally grow to $\infty$. In practice, we use an exponentially increasing schedule $\mu_k = \mu_0 \times a^k$ with small initial $\mu_0$ and appropriately chosen $a > 1$ for the $k$-th step of the LC. For most of compression schemes, we have developed robust estimates of $\mu_0$-values: for pruning see suppl.mat.\ of \cite{CarreirIdelbay18a}, for rank-selection see suppl.mat.\ of \cite{IdelbayCarreir20a}. For the value of $a$, we found the range of $[1.1\,\, 1.4]$ to be a good spot. 
\end{description}

\section{Conclusion}
The fields of machine learning and signal compression have developed independently for a long time: machine learning solves the problem of training a deep net to minimize a desired loss on a dataset, while signal compression solves the problem of optimally compressing a given signal. The LC algorithm allows us to seamlessly integrate the existing algorithms to train deep nets (L step) and algorithms to compress a signal (C step) by tapping on abundant literature in machine learning and signal compression fields. Based on this, we developed an extensible software framework that can easily plug in existing deep net training techniques with existing signal compression techniques and their combinations, in a flexible mix-and-match way. The source code is available online as an open-source project in Github.

\bibliographystyle{abbrvnat}

\end{document}